\documentclass[11pt,a4paper]{article}
\usepackage[hyperref]{naaclhlt2018}
\usepackage{times}
\usepackage{latexsym}
\usepackage{multirow}
\usepackage{multicol}
\usepackage{hhline}
\usepackage{graphicx}
\usepackage{subfigure}
\usepackage{url}
\usepackage{amssymb}
\usepackage{algorithm}
\usepackage{algorithmic}
\usepackage{caption}

\aclfinalcopy 
\def\aclpaperid{109} 

\title{Supervised and Unsupervised Transfer Learning for Question Answering}

\author{Yu-An Chung$^{1}$\quad Hung-Yi Lee$^{2}$\quad James Glass$^{1}$\\
  $^{1}$MIT Computer Science and Artificial Intelligence Laboratory, Cambridge, MA, USA\\
  {\tt \{andyyuan,glass\}@mit.edu}\\\\
  $^{2}$Department of Electrical Engineering, National Taiwan University, Taipei, Taiwan\\
  \tt hungyilee@ntu.edu.tw
}

\date{}

\begin{document}
\maketitle

\begin{abstract}
Although transfer learning has been shown to be successful for tasks like object and speech recognition, its applicability to question answering~(QA) has yet to be well-studied.
In this paper, we conduct extensive experiments to investigate the transferability of knowledge learned from a source QA dataset to a target dataset using two QA models.
The performance of both models on a TOEFL listening comprehension test~\citep{tseng2016towards} and MCTest~\citep{richardson2013mctest} is significantly improved via a simple transfer learning technique from MovieQA~\citep{tapaswi2016movieqa}.
In particular, one of the models achieves the state-of-the-art on all target datasets; for the TOEFL listening comprehension test, it outperforms the previous best model by 7\%.
Finally, we show that transfer learning is helpful even in unsupervised scenarios when correct answers for target QA dataset examples are not available.
\end{abstract}

\section{Introduction}
\subsection{Question Answering}
One of the most important characteristics of an intelligent system is to understand stories like humans do.
A story is a sequence of sentences, and can be in the form of plain text~\citep{trischler2017newsqa,rajpurkar2016squad,weston2016towards,yang2015wikiqa} or spoken content~\citep{tseng2016towards}, where the latter usually requires the spoken content to be first transcribed into text by automatic speech recognition~(ASR), and the model will subsequently process the ASR output.
To evaluate the extent of the model's understanding of the story, it is asked to answer questions about the story.
Such a task is referred to as question answering~(QA), and has been a long-standing yet challenging problem in natural language processing~(NLP).

Several QA scenarios and datasets have been introduced over the past few years.
These scenarios differ from each other in various ways, including the length of the story, the format of the answer, and the size of the training set.
In this work, we focus on context-aware multi-choice QA, where the answer to each question can be obtained by referring to its accompanying story, and each question comes with a set of answer choices with only one correct answer.
The answer choices are in the form of open, natural language sentences.
To correctly answer the question, the model is required to understand and reason about the relationship between the sentences in the story.

\subsection{Transfer Learning}
Transfer learning~\citep{pan2010survey} is a vital machine learning technique that aims to use the knowledge learned from one task and apply it to a different, but related, task in order to either reduce the necessary fine-tuning data size or improve performance. Transfer learning, also known as domain adaptation%
\footnote{In this paper, we do not distinguish conceptually between transfer learning and domain adaptation. A `domain' in the sense we use throughout this paper is defined by datasets.}, has achieved success in numerous domains such as computer vision~\citep{sharif2014cnn}, ASR~\citep{doulaty2015data,huang2013cross}, and NLP~\citep{zhang2017aspect,mou2016transferable}.
In computer vision, deep neural networks trained on a large-scale image classification dataset such as ImageNet~\citep{russakovsky2015imagenet} have proven to be excellent feature extractors for a broad range of visual tasks such as image captioning~\citep{lu2017knowing,karpathy2015deep,fang2015captions} and visual question answering~\citep{xu2016ask,fukui2016multimodal,yang2016stacked,antol2015vqa}, among others.
In NLP, transfer learning has also been successfully applied to tasks like sequence tagging~\citep{yang2017transfer}, syntactic parsing~\citep{mcclosky2010automatic} and named entity recognition~\citep{chiticariu2010domain}, among others.

\subsection{Transfer Learning for QA}
Although transfer learning has been successfully applied to various applications, its applicability to QA has yet to be well-studied.
In this paper, we tackle the TOEFL listening comprehension test~\citep{tseng2016towards} and MCTest~\citep{richardson2013mctest} with transfer learning from MovieQA~\citep{tapaswi2016movieqa} using two existing QA models.
Both models are pre-trained on MovieQA and then fine-tuned on each target dataset, so that their performance on the two target datasets are significantly improved.
In particular, one of the models achieves the state-of-the-art on all target datasets; for the TOEFL listening comprehension test, it outperforms the previous best model by 7\%.

Transfer learning without any labeled data from the target domain is referred to as unsupervised transfer learning.
Motivated by the success of unsupervised transfer learning for speaker adaptation~\citep{chen2011constrained,wallace2009study} and spoken document summarization~\citep{lee2013unsupervised}, we further investigate whether unsupervised transfer learning is feasible for QA.

Although not well studied in general, transfer Learning for QA has been explored recently.
To the best of our knowledge, \citet{kadlec2016from} is the first work that attempted to apply transfer learning for machine comprehension.
The authors showed only limited transfer between two QA tasks, but the transferred system was still significantly better than a random baseline.
\citet{wiese2017neural} tackled a more specific task of biomedical QA with transfer learning from a large-scale dataset.
The work most similar to ours is by~\citet{min2017question}, where the authors used a simple transfer learning technique and achieved significantly better performance.
However, none of these works study unsupervised transfer learning, which is especially crucial when the target dataset is small.
\citet{golub2017synthesis} proposed a two-stage synthesis network that can generate synthetic questions and answers to augment insufficient training data without annotations.
In this work, we aim to handle the case that the questions from the target domain are available.

\section{Task Descriptions and Approaches}
Among several existing QA settings, in this work we focus on multi-choice QA~(MCQA).
We are particularly interested in understanding whether a QA model can perform better on one MCQA dataset with knowledge transferred from another MCQA dataset.
In Section~\ref{sec:MCQA-formal}, we first formalize the task of MCQA.
We then describe the procedures for transfer learning from one dataset to another in Section~\ref{sec:transfer}.
We consider two kinds of settings for transfer learning in this paper, one is supervised and the other is unsupervised.

\subsection{Multi-Choices QA}
\label{sec:MCQA-formal}
In MCQA, the inputs to the model are a story, a question, and several answer choices.
The story, denoted by~$\mathbf{S}$, is a list of sentences, where each of the sentences is a sequence of words from a vocabulary set~$V$.
The question and each of the answer choices, denoted by~$\mathbf{Q}$ and~$\mathbf{C}$, are both single sentences also composed of words from~$V$.
The QA model aims to choose one correct answer from multiple answer choices based on the information provided in~$\mathbf{S}$ and~$\mathbf{Q}$.

\subsection{Transfer Learning}
\label{sec:transfer}
The procedure of transfer learning in this work is straightforward and includes two steps.
The first step is to pre-train the model on one MCQA dataset referred to as the \textbf{source} task, which usually contains abundant training data.
The second step is to fine-tune the same model on the other MCQA dataset, which is referred to as the \textbf{target} task, that we actually care about, but that usually contains much less training data.
The effectiveness of transfer learning is evaluated by the model's performance on the target task.

\subsubsection*{Supervised Transfer Learning}
\label{sec:supervised}
In supervised transfer learning, both the source and target datasets provide the correct answer to each question during pre-training and fine-tuning, and the QA model is guided by the correct answer to optimize its objective function in a supervised manner in both stages.

\subsubsection*{Unsupervised Transfer Learning}
\label{sec:unsupervised}
We also consider unsupervised transfer learning where the correct answer to each question in the target dataset is not available.
In other words, the entire process is supervised during pre-training, but unsupervised during fine-tuning.
A self-labeling technique inspired by~\citet{lee2013unsupervised,chen2011constrained,wallace2009study} is used during fine-tuning on the target dataset.
We present the proposed algorithm for unsupervised transfer learning in Algorithm~\ref{alg:unsupervised-TL}.
\begin{algorithm}[!htbp]
  \renewcommand{\algorithmicrequire}{\textbf{Input:}}
  \renewcommand{\algorithmicensure}{\textbf{Output:}}
  \caption{Unsupervised QA Transfer Learning}
  \label{alg:unsupervised-TL}
  \begin{algorithmic}[1]
    \REQUIRE
      Source dataset with correct answer to each question; Target dataset without any answer; Number of training epochs.
    \ENSURE
      Optimal QA model~$M^{*}$
    \STATE Pre-train QA model~$M$ on the source dataset.
    \REPEAT
      \STATE For each question in the target dataset, use~$M$ to predict its answer.
      \STATE For each question, assign the predicted answer to the question as the correct one.
      \STATE Fine-tune~$M$ on the target dataset as usual.
    \UNTIL{Reach the number of training epochs.}
  \end{algorithmic}
\end{algorithm}

\section{Datasets}
\label{sec:data}
\begin{table*}[ht]
\centering
\scriptsize
{\renewcommand{\arraystretch}{1.4}
\begin{tabular}{|c|c|c|c|}
\hline
\multirow{2}{*}{} & Source Dataset & \multicolumn{2}{c|}{Target Dataset} \\ \cline{2-4} 
 & \multicolumn{1}{c|}{MovieQA} & \multicolumn{1}{c|}{TOEFL} & \multicolumn{1}{c|}{MCTest} \\ \hline
$\mathbf{S}$ & \begin{tabular}[c]{@{}c@{}}After entering the boathouse, the trio witness\\ Voldemort telling Snape that the elder Wand\\ cannot serve Voldemort until Snape dies ...\\ Before dying, Snape tells Harry to take his\\ memories to the Pensieve ...\end{tabular} & \begin{tabular}[c]{@{}c@{}}I just wanted to take a few minutes to meet\\ with everyone to make sure your class\\ presentations for next week are all in order\\ and coming along well. And as you know,\\ you're supposed to report on some areas\\ of recent research on genetics ...\end{tabular} & \begin{tabular}[c]{@{}c@{}}James the Turtle was always getting in\\ trouble. Sometimes he'd reach into the\\ freezer and empty out all the food ...\\ Then he walked to the fast food restaurant\\ and ordered 15 bags of fries. He didn't\\ pay, and instead headed home ...\end{tabular} \\ \hline
$\mathbf{Q}$ & What does Snape tell Harry before he dies? & Why does the professor meet with the student? & What did James do after he ordered the fries? \\ \hline
$\mathbf{C}_{1}$ & To bury him in the forest & \begin{tabular}[c]{@{}c@{}}To find out if the student is interested\\ in taking part in a genetics project\end{tabular} & went to the grocery store \\ \hline
$\mathbf{C}_{2}$ & That he always respected him & \begin{tabular}[c]{@{}c@{}}To discuss the student's experiment\\ on the taste perception\end{tabular} & \textbf{went home without paying} \\ \hline
$\mathbf{C}_{3}$ & To remember to him for the good deeds & \begin{tabular}[c]{@{}c@{}}\textbf{To determine if the student has selected}\\ \textbf{an appropriate topic for his class project}\end{tabular} & ate them \\ \hline
$\mathbf{C}_{4}$ & \textbf{To take his memories to the Pensieve} & \begin{tabular}[c]{@{}c@{}}To explain what the student should\\ focus on for his class presentation\end{tabular} & made up his mind to be a better turtle \\ \hline
$\mathbf{C}_{5}$ & To write down his memories in a book &  &  \\ \hline
\end{tabular}
}
\caption{
  Example of the story-question-choices triplet from MovieQA, TOEFL listening comprehension test, and MCTest datasets.~$\mathbf{S}, \mathbf{Q}$, and~$\mathbf{C}_{i}$ denote the story, question, and one of the answer choices, respectively.
  For MovieQA, each question comes with five answer choices; and for TOEFL and MCTest, each question comes with only four answer choices.
  The correct answer is marked in bold.
}
\label{tab:data-example}
\end{table*}

We used MovieQA~\citep{tapaswi2016movieqa} as the source MCQA dataset, and TOEFL listening comprehension test~\citep{tseng2016towards} and MCTest~\citep{richardson2013mctest} as two separate target datasets.
Examples of the three datasets are shown in Table~\ref{tab:data-example}.

\paragraph{MovieQA} is a dataset that aims to evaluate automatic story comprehension from both video and text.
The dataset provides multiple sources of information such as plot synopses, scripts, subtitles, and video clips that can be used to infer answers.
We only used the plot synopses of the dataset, so our setting is the same as pure textual MCQA.
The dataset contains 9,848/1,958 train/dev examples; each question comes with a set of five possible answer choices with only one correct answer.

\paragraph{TOEFL listening comprehension test}
is a recently published, very challenging MCQA dataset that contains 717/124/122 train/dev/test examples.
It aims to test knowledge and skills of academic English for global English learners whose native languages are not English.
There are only four answer choices for each question.
The stories in this dataset are in audio form.
Each story comes with two transcripts: manual and ASR transcriptions, where the latter is obtained by running the CMU Sphinx recognizer~\citep{walker2004sphinx} on the original audio files.
We use TOEFL-manual and TOEFL-ASR to denote the two versions, respectively.
We highlight that the questions in this dataset are not easy because most of the answers cannot be found by simply matching the question and the choices without understanding the story.
For example, there are questions regarding the gist of the story or the conclusion for the conversation.

\paragraph{MCTest} is a collection of 660 elementary-level children's stories.
Each question comes with a set of four answer choices.
There are two variants in this dataset: MC160 and MC500.
The former contains 280/120/240 train/dev/test examples, while the latter contains 1,200/200/600 train/dev/test examples and is considered more difficult.

The two chosen target datasets are challenging because the stories and questions are complicated, and only small training sets are available.
Therefore, it is difficult to train statistical models on only their training sets because the small size limits the number of parameters in the models, and prevents learning any complex language concepts simultaneously with the capacity to answer questions.
We demonstrate that we can effectively overcome these difficulties via transfer learning in Section~\ref{sec:experiments}.

\section{QA Neural Network Models}
\label{sec:models}
Among numerous models proposed for multiple-choice QA~\citep{trischler2016parallel,fang2016hierarchical,tseng2016towards}, we adopt the End-to-End Memory Network~(MemN2N)%
\footnote{MemN2N was originally designed to output a single word within a fixed vocabulary set. To apply it to MCQA, some modification is needed. We describe the modifications in Section~\ref{sec:memn2n}.}~\citep{sukhbaatar2015end}
and Query-Based Attention CNN~(QACNN)%
\footnote{\url{https://github.com/chun5212021202/ACM-Net}}~\citep{liu2017query}, both open-sourced, to conduct the experiments.
Below we briefly introduce the two models in Section~\ref{sec:memn2n} and Section~\ref{sec:qacnn}, respectively.
For the details of the models, please refer to the original papers.

\subsection{End-to-End Memory Networks}
\label{sec:memn2n}
An End-to-End Memory Network~(MemN2N) first transforms~$\mathbf{Q}$ into a vector representation with an embedding layer~$B$.
At the same time, all sentences in~$\mathbf{S}$ are also transformed into two different sentence representations with two additional embedding layers~$A$ and~$C$. 
The first sentence representation is used in conjunction with the question representation to produce an attention-like mechanism that outputs the similarity between each sentence in~$\mathbf{S}$ and~$\mathbf{Q}$.
The similarity is then used to weight the second sentence representation.
We then obtain the sum of the question representation and the weighted sentence representations over~$\mathbf{S}$ as $\mathbf{Q}^\prime$.
In the original MemN2N, $\mathbf{Q}^\prime$ is decoded to provide the estimation of the probability of being an answer for each word within a fixed set.
The word with the highest probability is then selected as the answer.
However, in multiple-choice QA, $\mathbf{C}$ is in the form of open, natural language sentences instead of a single word.
Hence we modify MemN2N by adding an embedding layer~$F$ to encode~$\mathbf{C}$ as a vector representation $\mathbf{C}^\prime$ by averaging the embeddings of words in~$\mathbf{C}$. 
We then compute the similarity between  each choice representation $\mathbf{C}^\prime$ and $\mathbf{Q}^\prime$.
The choice $\mathbf{C}$ with the highest probability is then selected as the answer.

\subsection{Query-Based Attention CNN}
\begin{figure}[!htbp]
  \centering
  \includegraphics[scale=0.5]{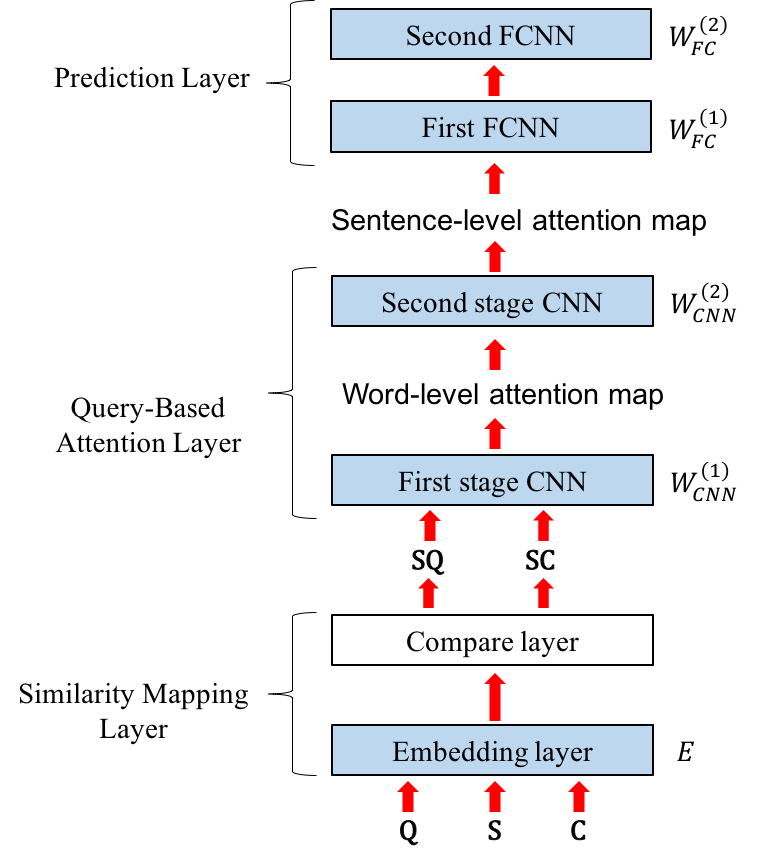}
  \caption{
    QACNN architecture overview.
    QACNN consists of a similarity mapping layer, a query-based attention layer, and a prediction layer.
    The two-stage attention mechanism takes place in the query-based attention layer, yielding word-level and sentence-level attention map, respectively. 
    The trainable parameters, including~$E, W_{CNN}^{(1)}, W_{CNN}^{(2)}, W_{FC}^{(1)},$ and~$W_{FC}^{(2)}$, are colored in light blue.
  }
  \label{fig:qacnn}
\end{figure}

A Query-Based Attention CNN~(QACNN) first uses an embedding layer~$E$ to transform~$\mathbf{S}, \mathbf{Q}$, and~$\mathbf{C}$ into a word embedding.
Then a compare layer generates a story-question similarity map~$\mathbf{SQ}$ and a story-choice similarity map~$\mathbf{SC}$.
The two similarity maps are then passed into a two-stage CNN architecture, where a question-based attention mechanism on the basis of~$\mathbf{SQ}$ is applied to each of the two stages.
The first stage CNN generates a word-level attention map for each sentence in~$\mathbf{S}$, which is then fed into the second stage CNN to generate a sentence-level attention map, and yield choice-answer features for each of the choices.
Finally, a classifier that consists of two fully-connected layers collects the information from every choice answer feature and outputs the most likely answer.
The trainable parameters are the embedding layer~$E$ that transforms $\mathbf{S}, \mathbf{Q},$ and~$\mathbf{C}$ into word embeddings, the two-stage CNN~$W_{CNN}^{(1)}$ and~$W_{CNN}^{(2)}$ that integrate information from the word to the sentence level, and from the sentence to the story level, and the two fully-connected layers~$W_{FC}^{(1)}$ and~$W_{FC}^{(2)}$ that make the final prediction.
We mention the trainable parameters here because in Section~\ref{sec:experiments} we will conduct experiments to analyze the transferability of the QACNN by fine-tuning some parameters while keeping others fixed.
Since QACNN is a newly proposed QA model has a relatively complex structure, we illustrate its architecture in Figure~\ref{fig:qacnn}, which is enough for understanding the rest of the paper.
Please refer to the original paper~\citep{liu2017query} for more details.
\label{sec:qacnn}

\section{Question Answering Experiments}
\label{sec:experiments}
\subsection{Training Details}
For pre-training MemN2N and QACNN on MovieQA, we followed the exact same procedure as in~\citet{tapaswi2016movieqa} and~\citet{liu2017query}, respectively.
Each model was trained on the training set of the MovieQA task and tuned on the dev set, and the best performing models on the dev set were later fine-tuned on the target dataset.
During fine-tuning, the model was also trained on the training set of target datasets and tuned on the dev set, and the performance on the testing set of the target datasets was reported as the final result.
We use accuracy as the performance measurement.

\subsection{Supervised Transfer Learning}
\subsubsection*{Experimental Results}
\begin{table}[!htbp]
\centering
\resizebox{\columnwidth}{!}{%
{\renewcommand{\arraystretch}{1.3} 
\setlength\tabcolsep{2.0pt} 
\begin{tabular}{|c|l|c|c|c|c|}
\hline
\multirow{2}{*}{Model}   & \multicolumn{1}{c|}{\multirow{2}{*}{Training}} & \multicolumn{2}{c|}{TOEFL} & \multicolumn{2}{c|}{MCTest} \\ \cline{3-6} 
						 &                           & manual               & ASR                  & MC160                 & MC500    			   \\ \hhline{|=|=|=|=|=|=|}
\multirow{5}{*}{QACNN}   &(a) Target Only            & 48.9                 & 47.5                 & 57.5                  & 56.4                  \\ \cline{2-6}
                         &(b) Source Only            & 51.2          	    & 49.2                 & 68.1                  & 61.5                  \\ \cline{2-6}
                         &(c) Source + Target        & 52.5          	    & 49.7                 & 72.1                  & 64.6                  \\ \cline{2-6}
                         &(d) Fine-tuned (1)         & 53.4 (4.5)           & 51.5 (4.0)           & \textbf{76.4} (18.9)  & 68.7 (12.3)           \\ \cline{2-6}
                         &(e) Fine-tuned (2)         & \textbf{56.1} (7.2)  & \textbf{55.3} (7.8)  & 73.8 (16.3)           & \textbf{72.3} (15.9)  \\ \cline{2-6}
                         &(f) Fine-tuned (all)       & 56.0 (7.1)           & 55.1 (7.6)           & 69.3 (11.8)           & 67.7 (11.3)           \\ \hhline{|=|=|=|=|=|=|}
\multirow{3}{*}{MemN2N}  &(g) Target Only            & 45.2                 & 44.4                 & 57.2                  & 53.6                  \\ \cline{2-6}
                         &(h) Source Only            & 43.7                 & 41.9                 & 56.8                  & 52.3                  \\ \cline{2-6}
                         &(i) Source + Target        & 46.8          	    & 45.7                 & 60.4                  & 56.9                  \\ \cline{2-6}
                         &(j) Fine-tuned             & 48.6 (3.4)           & 46.6 (2.2)           & 66.7 (9.5)            & 62.8 (9.2)            \\ \hhline{|=|=|=|=|=|=|}
\multicolumn{2}{|c|}{\citet{fang2016hierarchical}}   & 49.1                 & 48.8                 & -                     & -                     \\ \hline
\multicolumn{2}{|c|}{\citet{trischler2016parallel}}  & -                    & -                    & 74.6                  & 71.0                  \\ \hline
\multicolumn{2}{|c|}{\citet{wang2015machine}}        & -                    & -                    & 75.3                  & 69.9                  \\ \hline
\end{tabular}
}}
\caption{
  Results of transfer learning on the target datasets.
  The number in the parenthesis indicates the accuracy increased via transfer learning~(compared to rows (a) and (g)).
  The best performance for each target dataset is marked in bold.
  We also include the results of the previous best performing models on the target datasets in the last three rows.
}
\label{tab:transfer-learning}
\end{table}

Table~\ref{tab:transfer-learning} reports the results of our transfer learning on TOEFL-manual, TOEFL-ASR, MC160, and MC500, as well as the performance of the previous best models and several ablations that did not use pre-training or fine-tuning.
From Table~\ref{tab:transfer-learning}, we have the following observations.

\paragraph{Transfer learning helps.} 
Rows (a) and (g) show the respective results when the QACNN and MemN2N are trained directly on the target datasets without pre-training on MovieQA.
Rows (b) and (h) show results when the models are trained only on the MovieQA data.
Rows (c) and (i) show results when the models are trained on both MovieQA and each of the four target datasets, and tested on the respective target dataset.
We observe that the results achieved in (a), (b), (c), (g), (h), and (i) are worse than their fine-tuned counterparts (d), (e), (f), and (j).
Through transfer learning, both QACNN and MemN2N perform better on all the target datasets.
For example, QACNN only achieves 57.5\% accuracy on MC160 without pre-training on MovieQA, but the accuracy increases by 18.9\% with pre-training~(rows (d) vs. (a)).
In addition, with transfer learning, QACNN outperforms the previous best models on TOEFL-manual by 7\%, TOEFL-ASR~\citep{fang2016hierarchical} by 6.5\%, MC160~\citep{wang2015machine} by 1.1\%, and MC500~\citep{trischler2016parallel} by 1.3\%, and becomes the state-of-the-art on all target datasets.

\paragraph{Which QACNN parameters to transfer?} For the QACNN, the training parameters are~$E, W_{CNN}^{(1)}, W_{CNN}^{(2)}, W_{FC}^{(1)}$, and~$W_{FC}^{(2)}$~(Section~\ref{sec:qacnn}).
To better understand how transfer learning affects the performance of QACNN, we also report the results of keeping some parameters fixed and only fine-tuning other parameters.
We choose to fine-tune either only the last fully-connected layer~$W_{FC}^{(2)}$~while keeping other parameters fixed (row (d) in Table~\ref{tab:transfer-learning}), the last two fully-connected layers~$W_{FC}^{(1)}$ and~$W_{FC}^{(2)}$~(row (e)), and the entire QACNN~(row (f)).
For TOEFL-manual, TOEFL-ASR, and MC500, QACNN performs the best when only the last two fully-connected layers were fine-tuned; for MC160, it performs the best when only the last fully-connected layer was fine-tuned.
Note that for training the QACNN, we followed the same procedure as in~\citet{liu2017query}, whereby pre-trained GloVe word vectors~\citep{pennington2014glove} were used to initialize the embedding layer, which were not updated during training.
Thus, the embedding layer does not depend on the training set, and the effective vocabularies are the same.

\paragraph{Fine-tuning the entire model is not always best.} It is interesting to see that fine-tuning the entire QACNN doesn't necessarily produce the best result.
For MC500, the accuracy of QACNN drops by 4.6\% compared to just fine-tuning the last two fully-connected layers (rows (f) vs. (e)).
We conjecture that this is due to the amount of training data of the target datasets - when the training set of the target dataset is too small, fine-tuning all the parameters of a complex model like QACNN may result in overfitting.
This discovery aligns with other domains where transfer learning is well-studied such as object recognition~\citep{yosinski2014transferable}.

\paragraph{A large quantity of mismatched training examples is better than a small training set.} 
We expected to see that a MemN2N, when trained directly on the target dataset without pre-training on MovieQA, would outperform a MemN2N pre-trained on MovieQA without fine-tuning on the target dataset~(rows (g) vs. (h)), since the model is evaluated on the target dataset. 
However, for the QACNN this is surprisingly not the case - QACNN pre-trained on MovieQA without fine-tuning on the target dataset outperforms QACNN trained directly on the target dataset without pre-training on MovieQA~(rows (b) vs. (a)).
We attribute this to the limited size of the target dataset and the complex structure of the QACNN.

\subsubsection*{Varying the fine-tuning data size}
\begin{table}[!htbp]
\centering
\resizebox{\columnwidth}{!}{%
{\renewcommand{\arraystretch}{1.2}
\setlength\tabcolsep{3.0pt}
\begin{tabular}{|c|c|c|c|c|}
\hline
\multirow{2}{*}{\begin{tabular}[c]{@{}c@{}}Percentage of the target\\ dataset used for fine-tuning\end{tabular}} & \multicolumn{2}{c|}{TOEFL} & \multicolumn{2}{c|}{MCTest} \\ \cline{2-5} 
                                                                                                                 & manual       & ASR         & MC160        & MC500        \\ \hline
0                                                                                                                & 51.2     & 49.2    & 68.1      & 61.5     \\ \hline
25\%                                                                                                             & 53.9 (2.7)   & 52.3 (3.1)  & 70.3 (2.2)   & 65.6 (4.1)   \\ \hline
50\%                                                                                                             & 54.8 (0.9)   & 54.4 (2.1)  & 71.9 (1.6)   & 68.0 (2.4)   \\ \hline
75\%                                                                                                             & 55.3 (0.5)   & 54.8 (0.4)  & 72.5 (0.6)   & 71.1 (3.1)   \\ \hline
100\%                                                                                                            & 56.0 (0.7)   & 55.1 (0.3)  & 73.8 (1.3)   & 72.3 (1.2)   \\ \hline
\end{tabular}}
}
\caption{Results of varying sizes of the target datasets used for fine-tuning QACNN. The number in the parenthesis indicates the accuracy increases from using the previous percentage for fine-tuning to the current percentage.}
\label{tab:vary-target}
\end{table}
We conducted experiments to study the relationship between the amount of training data from the target dataset for fine-tuning the model and the performance.
We first pre-train the models on MovieQA, then vary the training data size of the target dataset used to fine-tune them.
Note that for QACNN, we only fine-tune the last two fully-connected layers instead of the entire model, since doing so usually produces the best performance according to Table~\ref{tab:transfer-learning}.
The results are shown in Table~\ref{tab:vary-target}%
\footnote{We only include the results of QACNN in Table~\ref{tab:vary-target}, but the results of MemN2N are very similar to QACNN.}.
As expected, the more training data is used for fine-tuning, the better the model's performance is.
We also observe that the extent of improvement from using 0\% to 25\% of target training data is consistently larger than using from 25\% to 50\%, 50\% to 75\%, and 75\% to 100\%.
Using the QACNN fine-tuned on TOEFL-manual as an example, the accuracy of the QACNN improves by 2.7\% when varying the training size from 0\% to 25\%, but only improves by 0.9\%, 0.5\%, and 0.7\% when varying the training size from 25\% to 50\%, 50\% to 75\%, and 75\% to 100\%, respectively.

\subsubsection*{Varying the pre-training data size}
\begin{table}[!htbp]
\centering
\resizebox{\columnwidth}{!}{%
{\renewcommand{\arraystretch}{1.2}
\setlength\tabcolsep{3.0pt}
\begin{tabular}{|c|c|c|c|c|}
\hline
\multirow{2}{*}{\begin{tabular}[c]{@{}c@{}}Percentage of MovieQA\\ used for pre-training\end{tabular}} & \multicolumn{2}{c|}{TOEFL} & \multicolumn{2}{c|}{MCTest} \\ \cline{2-5} 
                                                                                                                 & manual       & ASR         & MC160        & MC500        \\ \hline
0                                                                                                                & 48.9     & 47.6    & 57.5     & 56.4      \\ \hline
25\%                                                                                                             & 51.7 (2.8)   & 50.7 (3.1)  & 63.8 (6.3)   & 62.4 (6.0)   \\ \hline
50\%                                                                                                             & 53.5 (1.8)   & 52.3 (1.6)  & 67.3 (3.5)   & 66.7 (4.3)   \\ \hline
75\%                                                                                                             & 54.8 (1.3)   & 54.6 (2.3)  & 71.2 (3.9)   & 70.2 (3.5)   \\ \hline
100\%                                                                                                            & 56.0 (1.2)   & 55.1 (0.5)  & 73.8 (2.6)   & 72.3 (2.1)   \\ \hline
\end{tabular}}
}
\caption{Results of varying sizes of the MovieQA used for pre-training QACNN. The number in the parenthesis indicates the accuracy increases from using the previous percentage for pre-training to the current percentage.}
\label{tab:vary-source}
\end{table}
We also vary the size of MovieQA for pre-training to study how large the source dataset should be to make transfer learning feasible.
The results are shown in Table~\ref{tab:vary-source}.
We find that even a small amount of source data can help.
For example, by using only 25\% of MovieQA for pre-training, the accuracy increases 6.3\% on MC160.
This is because 25\% of MovieQA training set~(2,462 examples) is still much larger than the MC160 training set~(280 examples). 
As the size of the source dataset increases, the performance of QACNN continues to improve.

\subsubsection*{Analysis of the Questions Types}
\begin{figure}[!htbp]
  \centering
  \includegraphics[scale=0.4]{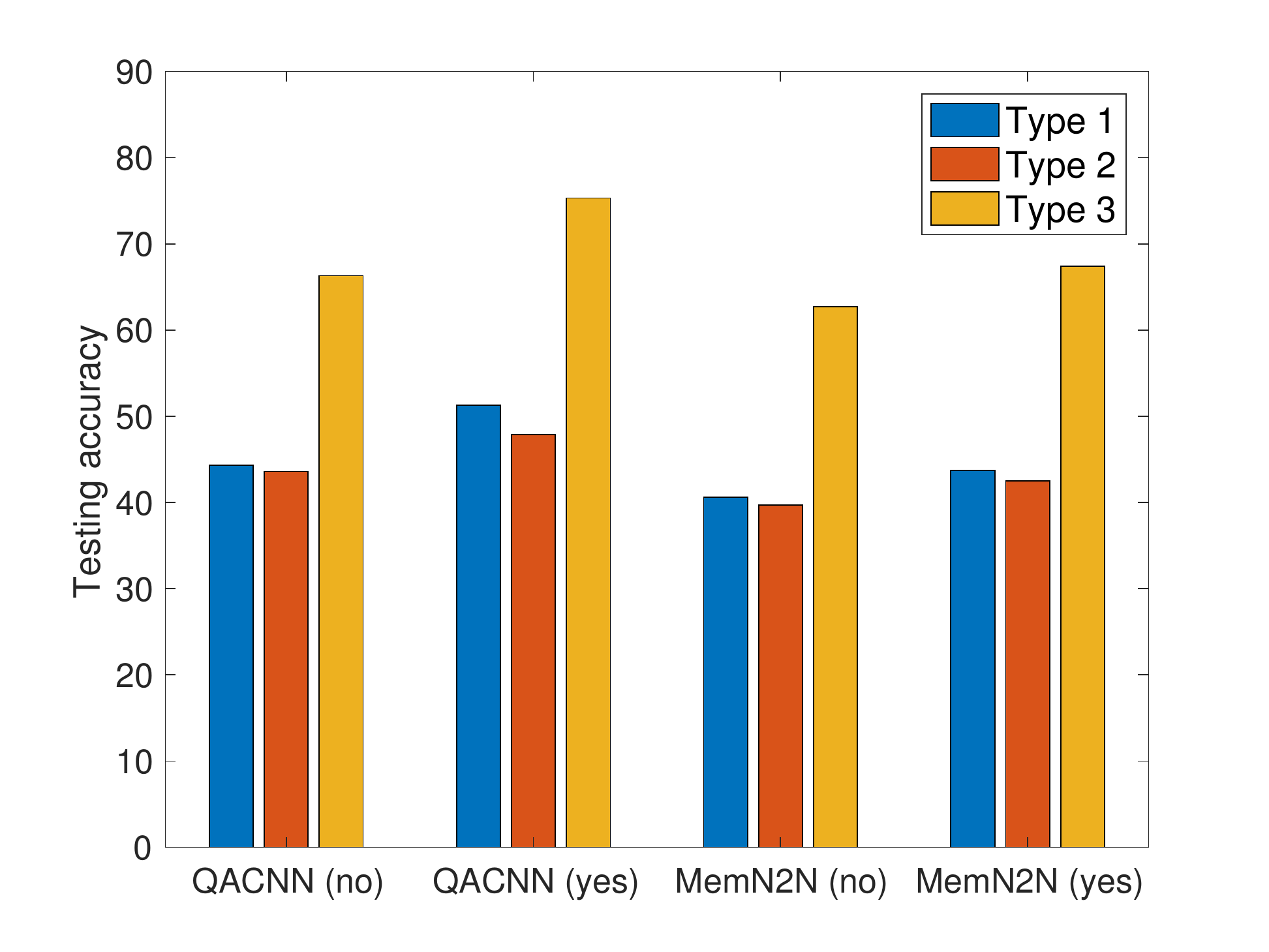}
  \caption{
    The performance of QACNN and MemN2N on different types of questions in TOEFL-manual with and without pre-training on MovieQA.
    `No' in the parenthesis indicates the models are not pre-trained, while `Yes' indicates the models are pre-trained on MovieQA.
  }
  \label{fig:different-types}
\end{figure}
We are interested in understanding what types of questions benefit the most from transfer learning.
According to the official guide to the TOEFL test, the questions in TOEFL can be divided into 3 types.
Type 1 questions are for basic comprehension of the story.
Type 2 questions go beyond basic comprehension, but test the understanding of the functions of utterances or the attitude the speaker expresses.
Type 3 questions further require the ability of making connections between different parts of the story, making inferences, drawing conclusions, or forming generalizations.
We used the split provided by~\citet{fang2016hierarchical}, which contains 70/18/34 Type 1/2/3 questions.
We compare the performance of the QACNN and MemN2N on different types of questions in TOEFL-manual with and without pre-training on MovieQA, and show the results in Figure~\ref{fig:different-types}.
From Figure~\ref{fig:different-types} we can observe that for both the QACNN and MemN2N, their performance on all three types of questions improves after pre-training, showing that the effectiveness of transfer learning is not limited to specific types of questions.

\subsection{Unsupervised Transfer Learning}
\begin{figure}[!htbp]
  \centering
  \subfigure[Results of TOEFL-manual and TOEFL-ASR]{
    \includegraphics[scale=0.19]{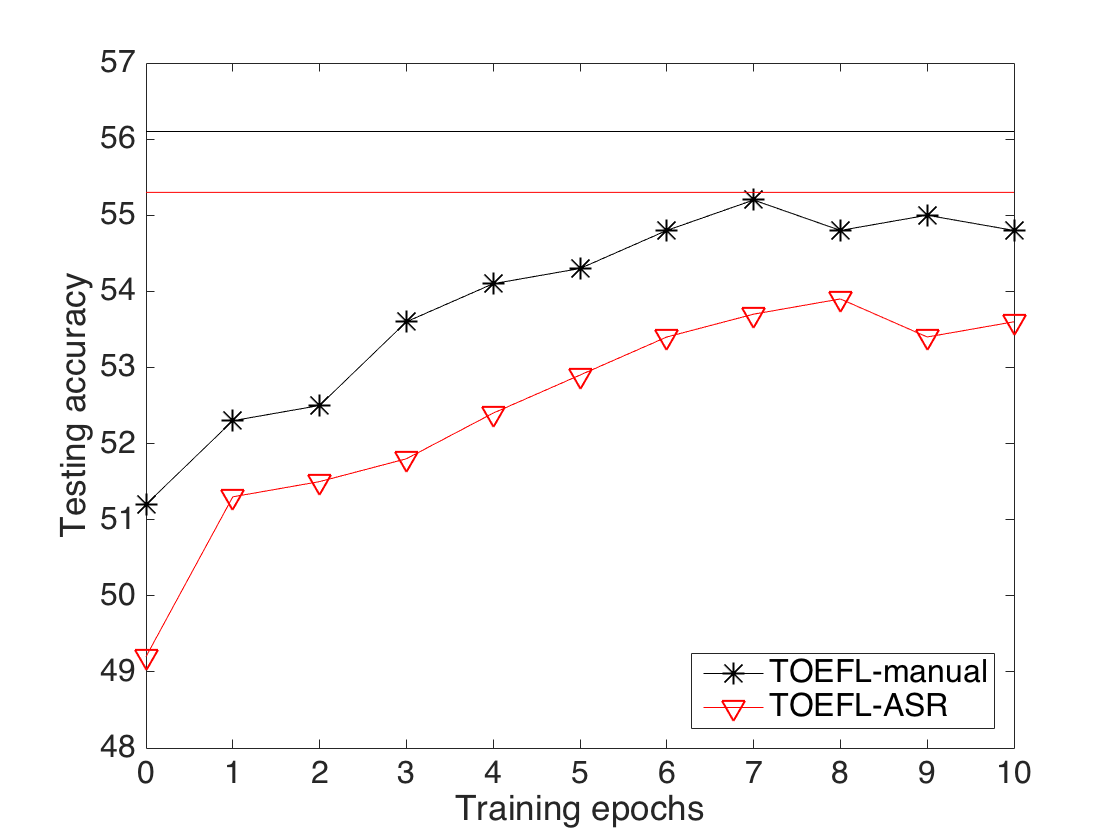}
    \label{fig:unsupervised-toefl}
  }\quad
  \subfigure[Results of MC160 and MC500]{
    \includegraphics[scale=0.19]{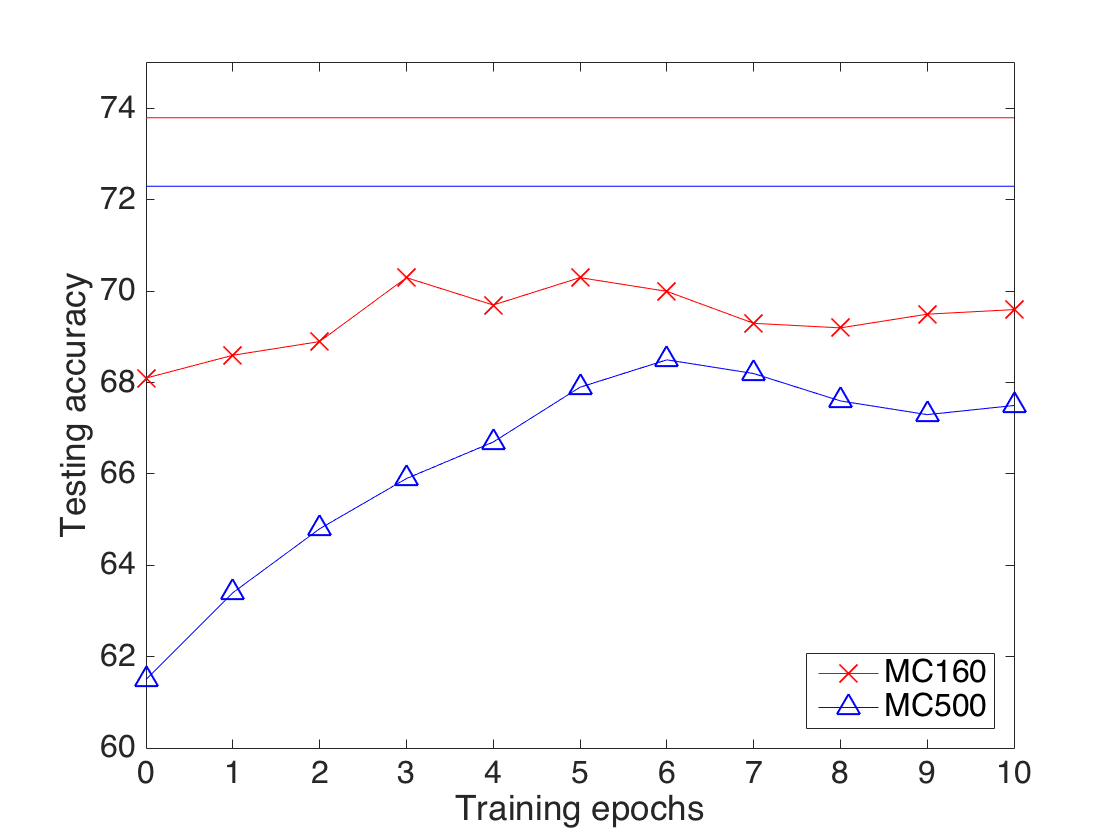}
    \label{fig:unsupervised-mctest}
  }
  \caption{
    The figures show the results of unsupervised transfer learning.
    The x-axis is the number of training epochs, and the y-axis is the corresponding testing accuracy on the target dataset.
    When training epoch = 0, the performance of QACNN is equivalent to row~(b) in Table~\ref{tab:transfer-learning}.
    The horizontal lines, where each line has the same color to its unsupervised counterpart, are the performances of QACNN with supervised transfer learning~(row~(e) in Table~\ref{tab:transfer-learning}), and are the upper-bounds for unsupervised transfer learning.
  }
\end{figure}
\begin{figure*}[!htbp]
  \centering
  \includegraphics[scale=0.57]{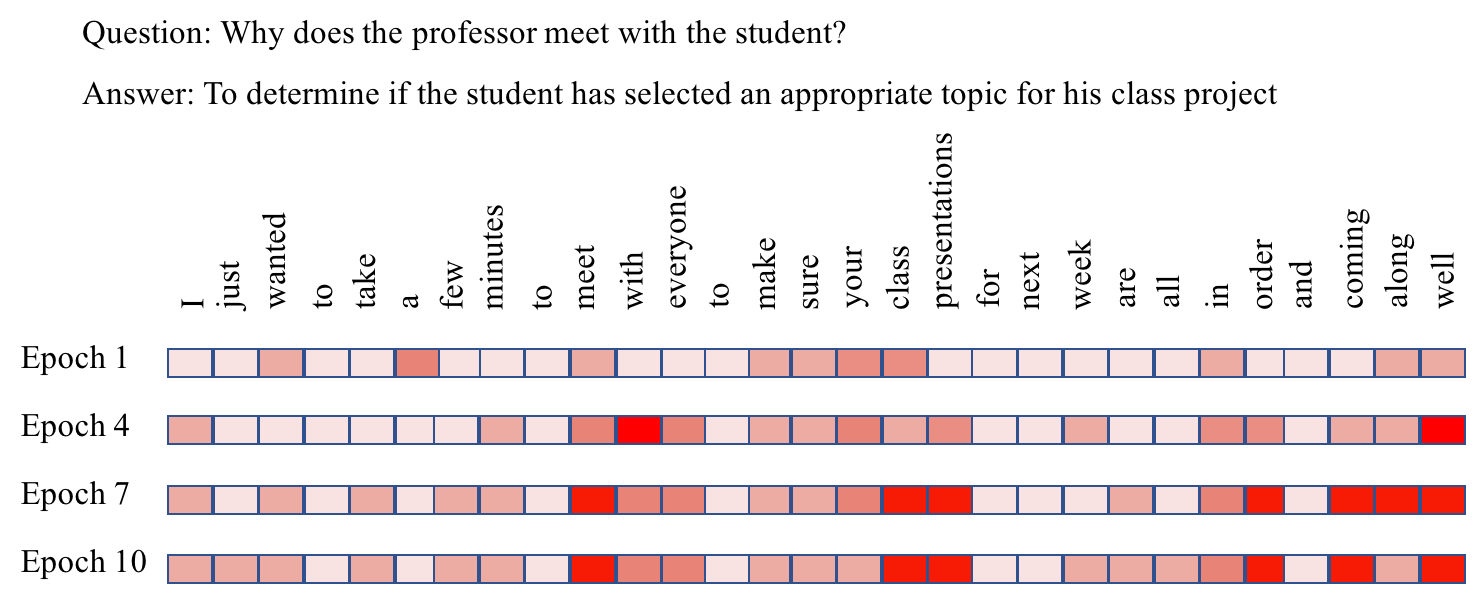}
  \caption{Visualization of the changes of the word-level attention map in the first stage CNN of QACNN in different training epochs. The more red, the more the QACNN views the word as a key feature. The input story-question-choices triplet is same as the one in Table~\ref{tab:data-example}.
  }
  \label{fig:attention-maps}
\end{figure*}
So far, we have studied the property of supervised transfer learning for QA, which means that during pre-training and fine-tuning, both the source and target datasets provide the correct answer for each question.
We now conduct unsupervised transfer learning experiments described in Section~\ref{sec:unsupervised}~(Algorithm~\ref{alg:unsupervised-TL}), where the answers to the questions in the target dataset are not available.
We used QACNN as the QA model and all the parameters~$(E, W_{CNN}^{(1)}, W_{CNN}^{(2)}, W_{FC}^{(1)},$ and~$W_{FC}^{(2)})$ were updated during fine-tuning in this experiment. 
Since the range of the testing accuracy of the TOEFL-series~(TOEFL-manual and TOEFL-ASR) is different from that of MCTest~(MC160 and MC500), their results are displayed separately in Figure~\ref{fig:unsupervised-toefl} and Figure~\ref{fig:unsupervised-mctest}, respectively.

\subsubsection*{Experimental Results}
From Figure~\ref{fig:unsupervised-toefl} and Figure~\ref{fig:unsupervised-mctest} we can observe that without ground truth in the target dataset for supervised fine-tuning, transfer learning from a source dataset can still improve the performance through a simple iterative self-labeling mechanism.
For TOEFL-manual and TOEFL-ASR, QACNN achieves the highest testing accuracy at Epoch 7 and 8, outperforming its counterpart without fine-tuning by approximately 4\% and 5\%, respectively.
For MC160 and MC500, the QACNN achieves the peak at Epoch 3 and 6, outperforming its counterpart without fine-tuning by about 2\% and 6\%, respectively.
The results also show that the performance of unsupervised transfer learning is still worse than supervised transfer learning, which is not surprising, but the effectiveness of unsupervised transfer learning when no ground truth labels are provided is validated.

\subsubsection*{Attention Maps Visualization}
To better understand the unsupervised transfer learning process of QACNN, we visualize the changes of the word-level attention map during training Epoch 1, 4, 7, and 10 in Figure~\ref{fig:attention-maps}.
We use the same question from TOEFL-manual as shown in Table~\ref{tab:data-example} as an example.
From Figure~\ref{fig:attention-maps} we can observe that as the training epochs increase, the QACNN focuses more on the context in the story that is related to the question and the correct answer choice.
For example, the correct answer is related to ``class project''. 
In Epoch 1 and 4, the model does not focus on the phrase ``class representation'', but the model attends on the phrase in Epoch 7 and 10. 
This demonstrates that even without ground truth, the iterative process in Algorithm~\ref{alg:unsupervised-TL} is still able to lead the QA model to gradually focus more on the important part of the story for answering the question.

\section{Conclusion and Future Work}
In this paper we demonstrate that a simple transfer learning technique can be very useful for the task of multi-choice question answering.
We use a QACNN and a MemN2N as QA models, with MovieQA as the source task and a TOEFL listening comprehension test and MCTest as the target tasks.
By pre-training on MovieQA, the performance of both models on the target datasets improves significantly.
The models also require much less training data from the target dataset to achieve similar performance to those without pre-training.
We also conduct experiments to study the influence of transfer learning on different types of questions, and show that the effectiveness of transfer learning is not limited to specific types of questions.
Finally, we show that by a simple iterative self-labeling technique, transfer learning is still useful, even when the correct answers for target QA dataset examples are not available, through quantitative results and visual analysis.

One area of future research will be generalizing the transfer learning results presented in this paper to other QA models and datasets.
In addition, since the original data format of the TOEFL listening comprehension test is audio instead of text, it is worth trying to initialize the embedding layer of the QACNN with semantic or acoustic word embeddings learned directly from speech~\citep{chung2018speech2vec,chung2017learning,chung2016audio} instead of those learned from text~\citep{mikolov2013distributed,pennington2014glove}.

\bibliography{mybib}
\bibliographystyle{acl_natbib}

\end{document}